# An Implementation of Back-Propagation Learning on GF11, a Large SIMD Parallel Computer

Michael Witbrock and Marco Zagha

December 1989

CMU-CS-89-208

School of Computer Science
Carnegie Mellon University
Pittsburgh, PA 15213

## Abstract

Current connectionist simulations require huge computational resources. We describe a neural network simulator for the IBM GF11, an experimental SIMD machine with 566 processors and a peak arithmetic performance of 11 Gigaflops. We present our parallel implementation of the backpropagation learning algorithm, techniques for increasing efficiency, performance measurements on the NETTALK text-to-speech benchmark, and a performance model for the simulator. Our simulator currently runs the back-propagation learning algorithm at 900 million connections per second, where each "connection per second" includes both a forward and backward pass. This figure was obtained on the machine when only 356 processors were working; with all 566 processors operational, our simulation will run at over one billion connections per second. We conclude that the GF11 is well-suited to neural network simulation, and we analyze our use of the machine to determine which features are the most important for high performance.

This research was performed at and supported by the IBM T.J. Watson Research Center, Yorktown Heights, NY 10598. The production of this Report was supported in part by Hughes Aircraft Corporation and National Science Foundation grant ECS-8716324.

## 1. Introduction

The recent development of several new and effective learning algorithms has inspired interest in applying neural networks to practical problems such as road following by autonomous vehicles[14], speech recognition[9], story understanding[12] and sonar target identification[7]. Many of these applications have been approached using variants of the Backpropagation learning algorithm[16]. Although this learning algorithm has been used to attack small problems with considerable success, the huge computational resources required have hindered attempts on large scale tasks.

One of the authors of this paper has recently been involved in an attempt, at Carnegie Mellon University, to apply backpropagation learning to the problem of speaker-independent continuous speech recognition [6]. Even for the relatively small digit recognition task initially selected[11], it has been necessary to train rather large recurrent nets[4] by making around 10,000 passes through very large amounts of data.[1] On the Convex C-1 used for these experiments, a typical training run took about a week. It became clear that the connectionist simulation tools available to us would make it very difficult to approach the ultimate goal of learning to transcribe continuously spoken general English. A simulator several orders of magnitude faster might permit a worthy attack.

The construction of such a simulator became feasible when the authors were offered the opportunity to write connectionist software for IBM's GF11 parallel supercomputer.

## 2. GF11 Architecture and Microcode Generation

GF11 [1] is an experimental parallel computer located at IBM's T.J. Watson Research Center at Yorktown Heights, New York. It is a SIMD (Single Instruction Multiple Data) machine composed of 566 processors interconnected through a Beněs Network[2] (See Figure 1). Each processor is capable of 20 million floating or fixed point operations per second, a rate which can be sustained during many kinds of calculations since intermediate results can be stored in a relatively large (256 word) register file. The registers can do up to four operations on every clock: write each of two operands to an ALU, read a result from an ALU, and write to or read from the interconnection network or static RAM (SRAM). Each processor has 16K words of static RAM, which, since it can be written or read on every clock, is effectively 4 times slower than the register file, and 512K words of dynamic RAM (DRAM), which is 4 times slower still. The Beněs network is capable of connecting the processors in arbitrary 1 to 1 permutations; 1024 such permutation patterns can be set up in the machine at once[2]. The whole machine has a peak arithmetic performance of 11.4 Gigaflops and contains a total of 1.14 Gbytes of semi-conductor memory.

Programs for GF11 consist of subroutines of sequential (non-branching) microcode residing on a single, system wide controller. All of GF11's 566 processors receive exactly the same instruction at exactly the same time. Typically each processor will apply the instructions to different data, and, because the processors have a table lookup facility, this data can reside at different local memory addresses on different processors. The controller is connected to an IBM RT Workstation, which schedules the execution of microcode subroutines residing in the controller, and which can read results from and write data to GF11's processors via the controller. The controller is capable of storing 512K microcode instructions, a length which which corresponds to about 1/40th of a second of GF11 run-time. Only very limited data dependent computation is possible in this code: *table lookup*, and *selection* of the source (data path) from which an operand is taken. Since neither branching nor looping is possible in the microcode, program flow control must be implemented by having the RT choose which stored microcode subroutine to run next. Since memory addresses contained in the microcode may be mapped onto different physical addresses (relocated) by the controller, one can loop through large arrays by repeatedly applying a single section of sequential microcode to different portions of processor memory.

---

[1]This long training period was necessary, even using the quickprop[5] variation of backprop which usually converges considerably faster than ordinary backprop.

[2]A variety of other connection topologies, including broadcasts and multicasts are also possible, but were relatively difficult to set up with the available software and were not needed for our purposes.

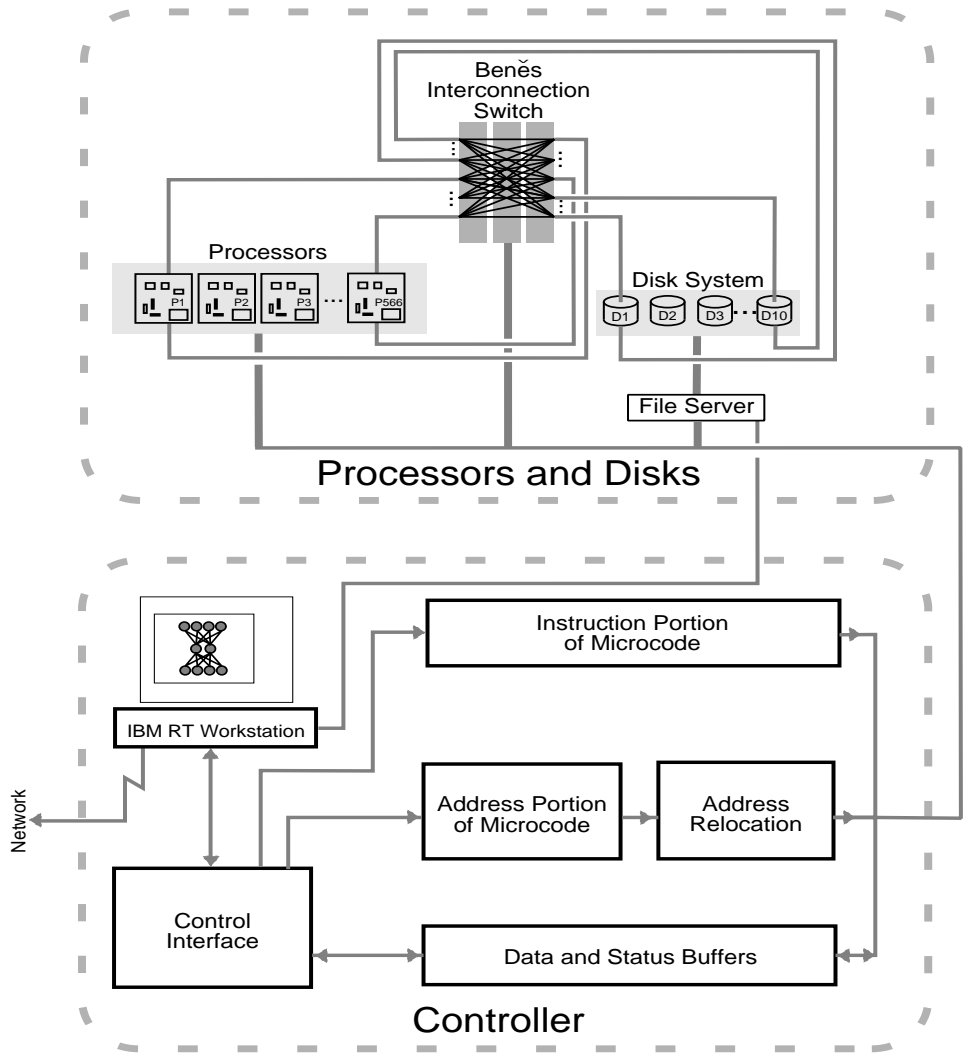

Figure 1: The Architecture of the GF11 Computer (after Beetem[1]).

Programming GF11 is straightforward. One can regard it as a vast floating point coprocessor attached to the RT. GF11 programs are programs written in a high level language[3] which run on the RT. At appropriate points in the program, functions are called to write data to, read data from, or execute a particular microcode subroutine on GF11. Microcode subroutines are, in turn, written as a series of calls to high-level language functions representing GF11 processor instructions such as floating add, write to memory, read from switch, etc. During a preliminary 'generation' execution of the program, these high level language subroutines are executed, and the GF11 operations required to perform their function are recorded. These operations are then passed through a scheduler and turned into a block of sequential microcode, which will execute on GF11 when the routine is called during a 'run' execution. Two advantages accrue from generating microcode by actually executing a high level program. The first advantage is that high level language constructs, such as loops and branches can be interspersed with the GF11 instructions, allowing the same block of high level language to generate many different microcode subroutines, depending on the state of program variables during 'generation'. This flexibility allowed our program to handle arbitrary network topologies very simply. Efficient, sequential microcode is generated by the program according to a topology file that it reads during 'generation'. In effect, our program *compiles* arbitrary feed-forward networks into GF11 microcode. The second advantage is that simulation of GF11 operation is virtually free. The routines representing GF11 operations are capable, depending on a mode switch, of outputting a GF11 microcode instruction, or of simulating execution of that instruction. We made a great deal of use of this simulation facility when developing and debugging our program.

## 3.  The Backpropagation Learning Algorithm

Backpropagation is a technique for training networks of simple neuron-like units connected by adjustable weights to perform arbitrary input/output mappings. Patterns are presented by setting the activation of a set of "input layer" units. Activation flows forward to units in subsequent layers via adjustable weights, eventually reaching the output layer. The activation of a unit is calculated by computing the weighted sum of the activations of the units connected to it and then applying a squashing, or logistic, function[4] to it. The object of learning is to minimize the value of some error metric[5] between the actual activations of the output units, and the values required by the desired input/output mapping. This is done by computing the effect of each weight on the error metric, and adjusting the weight in the direction of reduced error. Since both the weighted sum and the logistic function are differentiable, this can be done by computing partial derivatives of the error measure with respect to each weight, starting with the weights to the output units and working backwards.

---

[3]in the case of this simulator, the language used was a proprietary IBM language similar to PL1.

[4]Usually the sigmoid function, $\frac{1}{1+e^{-x}}$.

[5]Usually summed squared difference.

While in the "online" version of backprop, the required weight changes are applied as they are computed, in our implementation, the changes required to reduce the error measure for each input/output pattern are accumulated across all patterns and used to compute a net (or "pooled") weight update after all cases have been presented.

Pooled update differs from the online version of the backpropagation algorithm in the following way: if there are $n_p$ input-output pairs in the training sequence, online update allows $n_p$ weight changes to be applied, whereas the most extreme form of pooled update allows only 1. For some data-sets (the NETTALK training set among them), this is a decided disadvantage. Using online update, NETTALK can be learned in 10 complete passes through the training set[3], a performance unlikely to be matched by the 10 updates allowed by strict pooled update. It is, however, an open question whether pooled update is worse in general. For some tasks, it appears to work better; in some recent experiments performed by one of the authors (Witbrock) — training recurrent networks for speech recognition — online backprop, even with a very small learning rate, had a tendency to reach a plateau beyond which it could not successfully reduce the error. With pooled update, this effect was not noticed. Pooled update also allows one to store successive approximations to the slope of the weight space with respect to the training set, permitting the use of the Quickprop weight update rule. This rule converges considerably more quickly that the usual update rule for many problems [5, 10]. Finally, the apparent disadvantage of pooled update is reduced when observes that update doesn't have to be pooled over all $n_p$ patterns. In fact, weights can be updated after one case has been processed on each processor.

## 4. Simulator Implementation

### 4.1. Parallelizing Backprop for GF11

There are two obvious approaches to parallelizing backprop. In one approach, one divides the network, distributing 'neural processing units' — weights, units, or layers of units — across physical processors, and communicates activation levels between processors. In the other approach one can parallelize across training cases, having each processor simulate identical networks, but apply them to different subsets of training examples, communicating collected weight changes at the end of each training epoch (i.e. after a single presentation of each training example). Both approaches have been used in previous simulators. Blelloch and Rosenberg, in their simulator for the Connection Machine, mapped both individual weights and units to processors. The simulator on Warp initially mapped subsets of units with their corresponding weights to physical processors, but was later changed to the more efficient (on the Warp) case-based parallelism. Kevin Lang's [personal communication] simulator for the Convex C-1 Vector processor, and its descendants[6] all succeed in making backprop vectorizable by parallelizing across training cases.

Dividing units across processors is a technique which appears to be best suited to MIMD machines with very fast communications — MIMD, because different units in a neural network

---

[6]Written by Franzini and Witbrock.

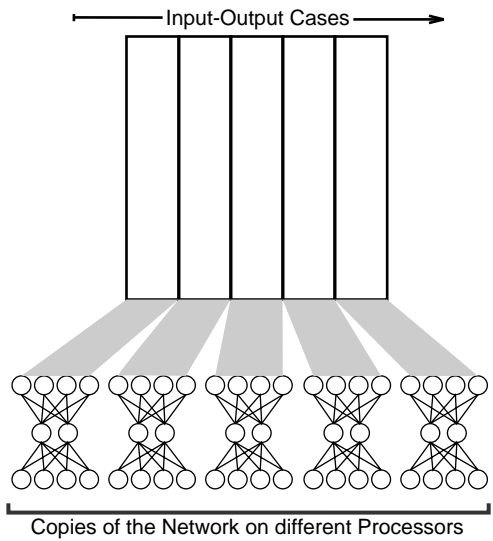

Input-Output Cases

Copies of the Network on different Processors

Figure 2: Parallelizing backprop for GF11

have different patterns of connection, and very fast communication because activation levels must be passed between processors for each pattern presentation. If both units and weights are divided across processors, SIMD suffices, since all weights (and all units) are essentially the same, but even more communication is necessary[7]. For SIMD machines with moderate numbers of processors and sufficient memory on each processor, the technique of having each processor run the same network (and hence the same code) over a different subset of cases maps more neatly onto the architecture. This latter technique is the one that we used when designing our simulator.

The ability to parallelize backprop across cases assumes that the changes to weights due to each input/output pattern are independent. That is, we must be able to satisfy the condition that the weight changes from the each of patterns in the training set can be applied in any order and yield the same result. This condition is clearly not satisfied by the canonical version of backprop, in which weight changes are calculated and *applied* after each pattern presentation. Instead, we use the "pooled update" technique where weight changes are summed across all input/output cases, and the net change is applied after the entire training set[8] has been presented.

When the algorithm is parallelized this way on GF11, each processor stores the input/output cases it is responsible for (See figure 2). It also has its own copy of the entire network (both units and weights), and storage for accumulating the total weight change due to its cases. After all the processors have finished calculating weight changes for all their cases, they share their accumulated weight changes with the other processors over the interconnection switch. They then update their local weights with the resulting total weight change over all cases.

---

[7]In fact, Blelloch and Rosenberg report that communication speed was the primary limitation on the speed of their Connection Machine simulator.

[8]Or a large enough subset of it to parallelize over.

## 4.2. Simulator Structure

Dividing the functionality of the simulator into blocks of microcode is straightforward, although there are a few software and hardware imposed constraints. The current software restricts the size of microcode blocks to 20K lines. The hardware design forces the programmer to work within the memory limits of the machine — in particular, the 256 word register file and the total size of microcode memory (512K lines).

We process weights by subsets of layers, or *bundles*, typically on the order of 1000 weights. This enables us to keep all the inputs and outputs of units in registers.

To save microcode, we use the same microcode for all training examples and simply set a pointer to the current example in the controller. Similarly, when updating weights, we repeatedly set pointers to the weights and weight changes and call a routine to update a small group of weights.

## 4.3. Processor Dependent Computation

Even when parallelization over cases is used to fit the algorithm to its SIMD architecture, there is still a problem with implementing backprop on the GF11. The problem arises because of the necessity of computing the sigmoid function $\frac{1}{1+e^{-x}}$, on processors which can not perform division, let alone exponentiation, of floating point numbers. They can, however, do addition, multiplication, and a few other IEEE floating point operations including, perhaps most importantly, extracting the integer part of a floating point number ($\lfloor x \rfloor$). They can also do table lookup within SRAM, and can select the source of operands depending on condition codes set as a result of previous operations (the "select" operation). Two methods of computing the sigmoid function were tried, using both forms of processor dependent computation available.

The first approach involved replacing the sigmoid function with $\frac{1}{1+2^{-x}}$, or equivalently,[9] $\frac{2^x}{2^x+1}$. To help compute $2^x$, we use the identity $2^x = 2^{\frac{x}{2}^2}$ and a polynomial approximation for $2^x$ accurate in the range $[0.0; 0.5]$. More precisely, we compute $2^{\lfloor x \rfloor} 2^{\frac{x-\lfloor x \rfloor}{2}^2}$, where $2^{\lfloor x \rfloor}$ is computed by table lookup. This reformulation reduces the required operations to addition, multiplication and round down, all of which are provided by the hardware, and taking the reciprocal of a number, which was available as an accurate library routine.

The second approach to computing the sigmoid function involved using the "select" operation (data dependent data path selection) to ensure that values of $x$ were in the range $[-15; 15]$. Values outside of this range were mapped to the extreme values of the sigmoid function. Numbers within the range were remapped, by addition and multiplication, as $x_1$ over a range of $[0; 255]$. The integer part $\lfloor x_1 \rfloor$ of these numbers was used to do table lookup into a precomputed table of $\frac{1}{1+e^{-x}}$; the fractional part of the number, $x_1 - \lfloor x_1 \rfloor$, was used to perform linear interpolation between successive entries in the lookup table. Using this combination of table lookup and interpolation,

---

[9] any differentiable monotonic function over the reals is usable

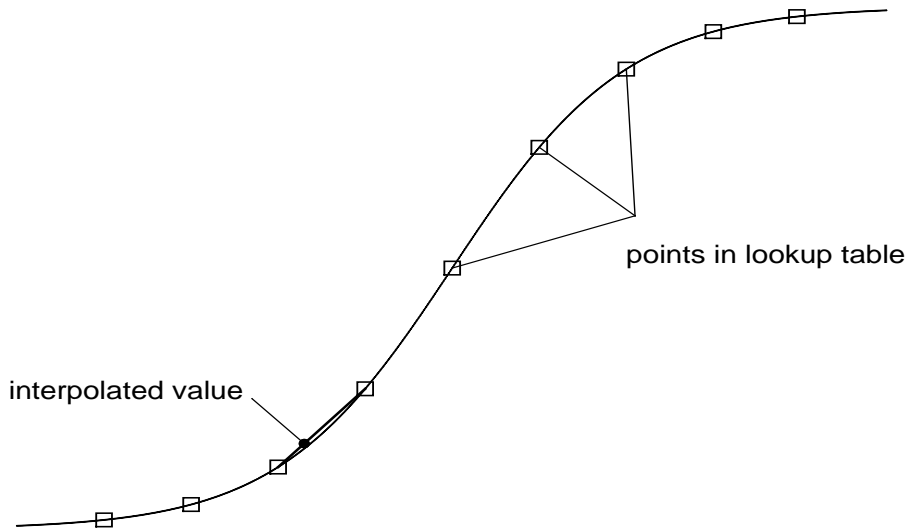

points in lookup table

interpolated value

Figure 3: Computing sigmoid by table lookup. Actual table contained 256 points.

the sigmoid function could be computed with approximately five decimal places of accuracy over its entire domain. See figure 3.

There were two other, less important, pieces of code where the "select" operation was used. One use was to replace outlandish weight changes by zero, so that the hardware errors which were frequent while we were developing our code would not render the algorithm unworkable (see section 8 below).

The other use of data dependent data path selection was in an (unfinished) implementation of Fahlman's *quickprop* weight update rule[5], which includes a number of calculation steps which are dependent on the local curvature of the error surface in weight space.

### 4.4.   Processor Communication

### 4.4.1.   Summing Weight Changes in a Ring

In our initial approach, we used one switch configuration to connect $P$ processors in a ring and sum each weight change in $P$ steps.

On the first step, each processor initializes *Sum* to zero and loads its weight change into the variable *Neighbor*. Then for remaining $P-1$ steps, each processor sends *Neighbor* to the processor on its right, receives a new value of *Neighbor* from the processor on its left, and adds *Neighbor* to *Sum*. At the end of $P$ steps each processor has the same value for the total weight change in *Sum* (See figure 4).

This algorithm has a time complexity of $O(NPROCS)$ per weight.

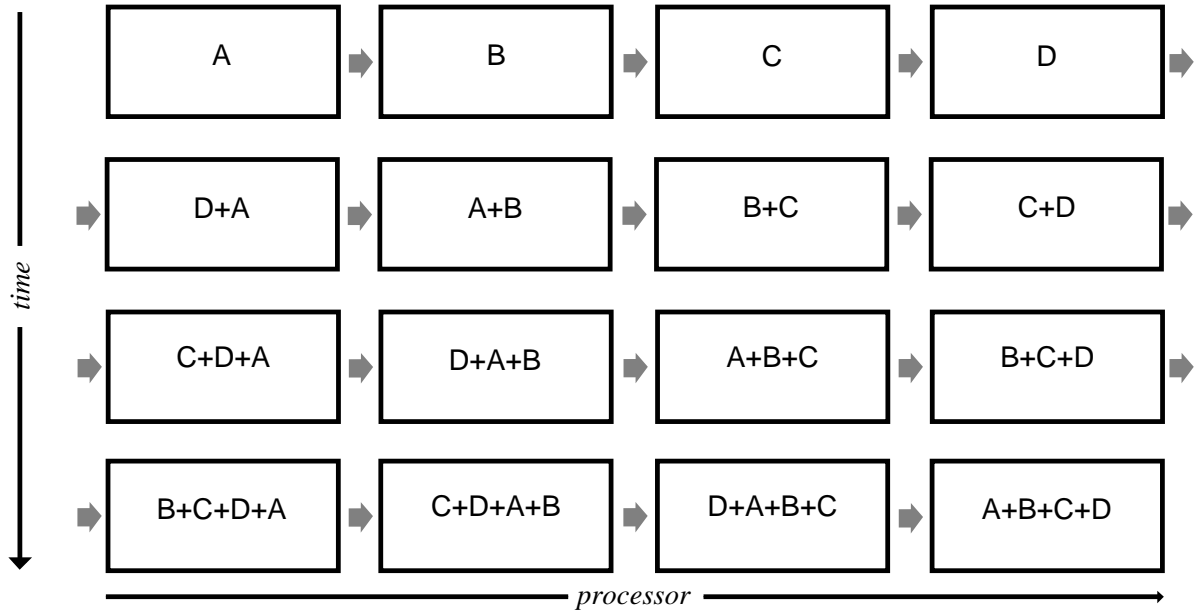

Figure 4: Summing Weight Changes with Processors Configured in a Ring

### 4.4.2. Summing Weight Changes in a Tree

GF11's powerful communication facility allows a more efficient approach to summing weight changes. We use several switch configurations to sum weight changes in a binary tree using a standard data-parallel algorithm[8].

On step $i$ (starting at zero), processor $P$ sends its weight change to processor $(P + 2^i)$ mod $NPROCS$ and adds the weight change it received to its weight change. After $\log_2 NPROCS$ steps, the total weight change in each processor contains the sum of the individual weight changes. See figure 5.

This algorithm has a time complexity of $O(\log_2 NPROCS)$ per weight.

If the number of processors is not a power of 2, we generate two additional communication instructions. Call each processor $\{P_n : n > 2^{\lfloor \log_2 NPROCS \rfloor}\}$ a "leftover" processor. Each leftover processor $P_n$ sends its weight change to its "buddy", processor number $n - 2^{\lfloor \log_2 NPROCS \rfloor}$, which adds the value to its weight change. Then logarithmic summing proceeds as described above. Finally, each leftover processor receives the final total weight change from its buddy. During the two extra steps, processors not involved in the communication mask out their operations using condition codes.

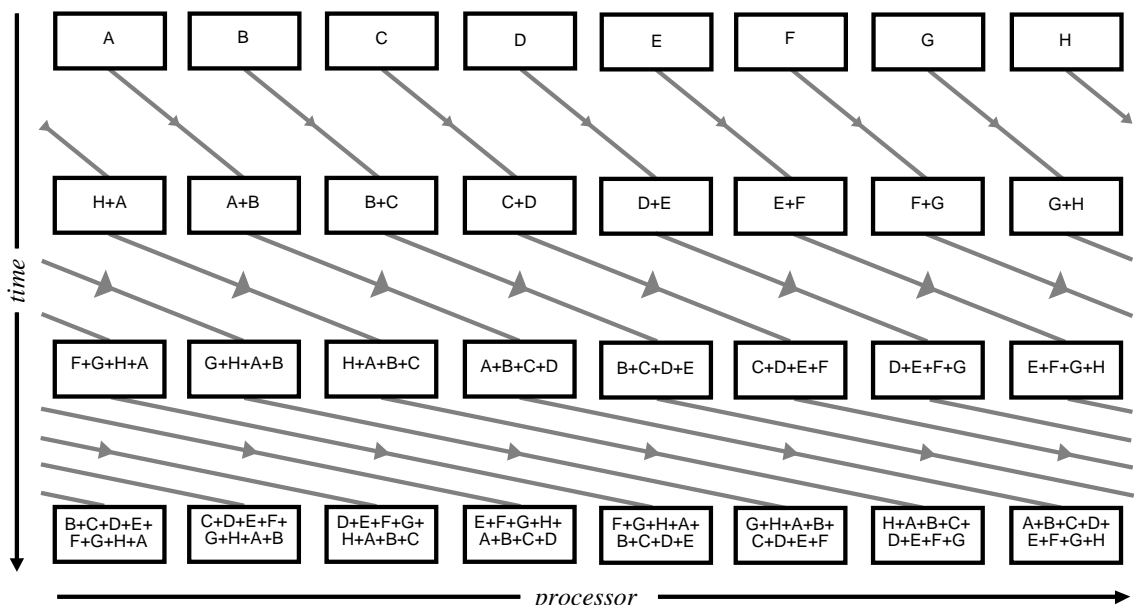

Figure 5: Summing Weight Changes with Processors Configured in a Tree

## 5. Simulating Larger Networks

As described so far, our implementation could only accommodate networks with a few thousand weights within the 16K word SRAM. Training examples can be kept in DRAM with trivial changes to the code and a negligible performance penalty. However, without further changes to the structure or the program, keeping weights in DRAM creates a fundamental efficiency problem: transfers to and from DRAM would take over twice as long as the floating point computation. A transfer to or from DRAM can be started at most once per 4 cycles. However, in the forward pass only two floating point operations per weight are required, yielding a maximum efficiency of 50%. In the backward pass, a weight change and a weight must be loaded, add and weight change must be stored, requiring 3 transfers per 4 floating point operations, yielding only 33% efficiency.

We adopt the following approach to obtain locality of reference to the weights and weight changes. Instead of doing independent forward and backward passes for each case, we move a bundle of weights and weight changes into an SRAM cache, process *several* training examples on those weights, and then move the weights values back to DRAM.

The original structure of the simulator was the following:

```
For each training example
  For each bundle of weights
    Sweep forward on bundle
  For each bundle of weights (in reverse order)
    Sweep backward on bundle
```

The new structure that uses DRAM:

```
For each group of training examples
  For each bundle of weights
    Move weights to a cache in SRAM
    For each training example in group
     Sweep forward on bundle
  For each bundle of weights (in reverse order)
    Move weights and weight changes to a cache in SRAM
    For each training example in group
      Sweep backward on bundle
    Move weight changes from cache to DRAM
```

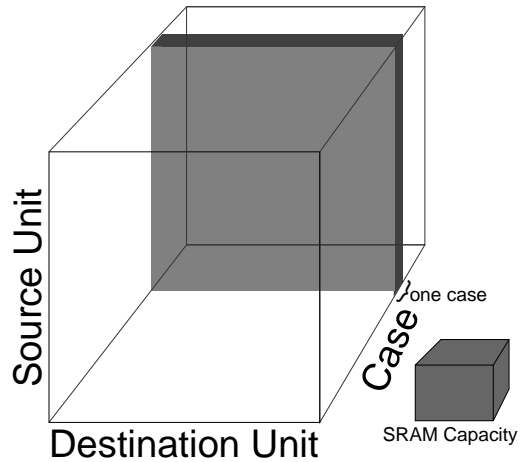

Figure 6: Processing all the Weights for a Single Case.

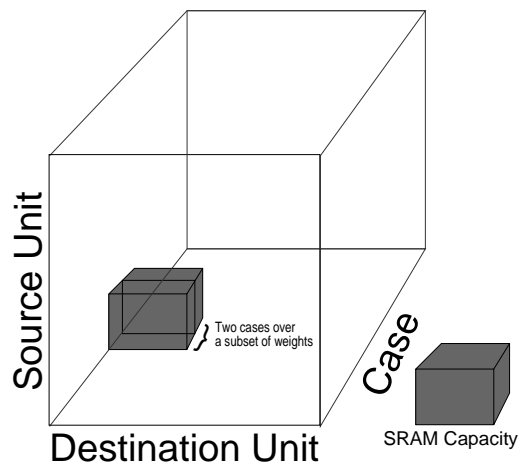

Figure 7: Processing a Subset of the Weights for Several Cases at Once.

As another way of understanding this change, the loop structure can be viewed as a traversal of points on a 3-dimensional rectangular grid. The points on the grid represent weights (or weight changes) and the axes of the grid are input unit index, output unit index, and training example index. In our original approach (see figure 6), we traverse one slice of the grid at once — that is we process one case at a time. We were forced to modify this algorithm because the memory required for this slice exceeds the capacity of free memory in SRAM. In the improved approach we traverse a sub-grid (See figure 7) which corresponds to several training examples processed on a subset of the weights. Since the weights are the same for each training example, we reduce transfers to and from DRAM.

## 6.   Simulator Performance

### 6.1.   Performance Model

Developing a performance model yields several important benefits. Execution times for different topologies and training set sizes can be estimated without having to execute a program on GF11. The performance model can also approximate the optimal number of processors to use for a given problem. More importantly, the model makes analyzing the role of the simulator components much easier. Bottlenecks are revealed, and the importance of specific machine features can be determined.

Several assumptions are implicit in the performance model. We neglect computation on units, such as the sigmoid calculation, since there are typically many more weights than units. We also assume that computation during the forward and backward passes fills the processor pipeline — this amounts to assuming more than 12 units per layer.

*Let*

| | |
|---|---|
| $W$ = total number of weights | $G$ = number of groups of training cases |
| $W_i$ = number weights in the input layer | $U$ = number of units |
| $B$ = maximum number of weights in a bundle | $M$ = number of free words in SRAM |
| $C$ = number of training cases | $P$ = number of processors |
| | $T$ = time per epoch (in cycles) |

In the forward pass, one add and one multiply per weight are executed:

$$\text{Cycles forward/case} = 2W$$

In the backward pass, two adds and two multiplies per weight are executed, except for connections from input units which execute one and and one multiply:

$$\text{Cycles backward/case} = 2W_i + 4(W - W_i)$$

Transfers to or from DRAM can only be started once every 4 cycles. In the forward pass, each weight is transferred to SRAM once per group of training examples. In the backward pass, a weight change is loaded and stored once per group of examples, and weights not in the input layer are loaded once:

$$\text{Cycles forward weight transferring/group} = 4W$$

$$\text{Cycles backward transferring weight changes/group} = (4)(2)W$$

$$\text{Cycles backward transferring weight/group} = 4(W - W_i)$$

Communications over the switch also can only be started once every 4 cycles.

When summing in a tree:[10]

$$\text{Cycles per update} = 4W\lceil \log_2 P \rceil$$

When summing in a ring:

$$\text{Cycles per update} = 4WP$$

After allocating space for the sigmoid lookup tables and a few other constants, the amount of free SRAM is approximately:

$$M = 15K$$

The weight value and weight change caches use $2B$ words of memory, and two words per unit are used to hold unit activation and error levels. Thus, the size of a group of cases that can be processed entirely in SRAM is:

$$\frac{M - 2B}{2U}$$

The number of groups of cases is:

$$G = C/P \times (2U/(M - 2B))$$

Total time per epoch in cycles (when summing in a tree):

$$T = 2W \times C/P + (2W_i + 4(W - W_i)) \times C/P + (12W + 4(W - W_i)) \times G + 4P \times W$$

Because each processor executes 20 million instructions per second, millions of connections per second (MCPS) is simply,

$$20/T$$

To find the optimal number of processors when summing in a tree we differentiate time with respect to the number of processors, equate the result to zero, solve, and simplify:

$$P = \frac{C}{4W} \times ((6W - 2W_i) + \frac{2U}{M - 2B} \times (16W - 4W_i))$$

When summing in a ring, the optimal number of processors is simply the square root of the above quantity.

---

[10]if the number of processors is not a power of 2, there is one extra communication not counted in the model

## 6.2.   Optimizing Microcode

### 6.2.1.   Optimization Techniques

Typically, the first attempt at implementing a microcode routine resulted in code running at less than 50% of optimal performance. In this section, we describe techniques for improving microcode efficiency. These techniques follow from two general principles of code optimization: breaking computation into carefully sized pieces to accomodate multi-level memory hierarchies, and reordering independent instructions to improve pipeline scheduling.

Producing efficient microcode is difficult for a number of reasons. The processor pipeline is 25 deep — that is, the result of a floating point operation does not appear for 25 cycles. In addition, several operations may execute in one cycle. Although pipeline scheduling is not the programmer's responsibility, the scheduling process is not completely invisible. The programmer often has to know what performance constraints the machine imposes and a few details of the scheduler implementation.

The microcode scheduler rearranges instruction order but is constrained by the programmer's approach to data movement. Thus, managing the register/SRAM/DRAM hierarchy becomes the key to producing efficient microcode. In section 5 we discussed changes to the program structure to minimize DRAM traffic. To minimize SRAM traffic, we keep net inputs in registers during the forward pass rather than reloading the values to increment them. Similarly, during the backward pass, error into a unit is kept in registers. Using weight bundles instead of entire weight layers reduces the number of units used in a block of microcode and enables this strategy to work within the 256 register limit.

Analyzing the output of the microcode scheduler is the next step after developing a reasonable approach to memory management. This phase can be tedious, but it often reveals problems that can fixed by changing the order of instructions or by inserting scheduler directives. For example, the routines that sum weight changes must loop over both communication steps and indices of weight changes. By simply interchanging the loops to sum all weight changes simultaneously, pipeline scheduling improves significantly. As another example, we improved the routine that sums weight changes in a ring by having it circulate weight increments rather than partial sums of weight increments. This change did not reduce the number of source instructions, but it did cut the number of dependent instructions in the critical path in half, since additions could be done in parallel with switch communication.

Inefficient scheduling in the backward pass motivated one of the most interesting optimizations. In both the forward and backward passes, we ordered computation on weights by the unit on the input side of the weight. This turns out to work well for the forward pass, but not in the backward pass because for a $M$ by $N$ bundle of weights, the last $M$ operations all increment the same register (the error from the last source unit). Since each operation must be separated by 25 cycles, this led to sub-optimal code. To improve efficiency, at microcode generation time, we simply create an ordering of weight indices for the backward pass sorted by output unit index. Then accesses to each of the $M$ input units are separated by at least $2N$ operations, which produces near optimal code for $N$ greater than half the pipeline depth (i.e., $N > 12$).

Interleaving code that is logically independent is the most difficult form of optimization, but is necessary for operations with a low throughput such as switch communication and transfers to and from DRAM. For example the routine responsible for updating weights performs several functions: it moves weights, weight changes, and weight changes from the previous epoch from DRAM to a cache in SRAM, sums weight deltas over the switch, calculates weight changes using the current value of the learning rate and momentum, updates the weights in the SRAM cache, moves the new weights back to DRAM, and zeroes weight changes in DRAM for the next epoch.

Finally, because the DRAM is interleaved, accesses should be sequential whenever possible.

### 6.2.2. Microcode Efficiency

In this section we report the efficiency of microcode routines generated for networks with at least 12 units per layer. We defer discussion of small networks until section 10.3.

In the forward pass, 97–99% of all cycles execute an add or multiply. Memory transfers execute in parallel with floating point computation.

In the backward pass, 85–99% of all cycles execute an add or multiply. This routine is slightly less efficient than the forward pass because of additional memory traffic — each weight requires both a load and store instead of just a load.

Routines calculating the sigmoid and back error for units execute an add or multiply on about 40–70% of all cycles, which is fast enough to make the execution time negligible.

Analyzing the performance for weight updating code is more complicated because we overlap DRAM transfers and computation with communication as described in the previous section. For simplicity, we adopt a pessimistic efficiency rating base on the number of switch communications only. Since each switch communication operation takes four cycles, we define the optimal number of cycles as four times the number of communications. When summing in a ring, we achieve 98% of optimal performance. DRAM transfers, weight update with momentum, and zeroing of weight changes for next epoch come for free in empty slots during communication. When summing in a tree, efficiency is approximately 60% of optimal performance because computation can not be fully overlapped with communication. Despite its lower execution *efficiency*, tree update takes considerably less time than ring update when more than 4 processors are used.

### 6.3. Performance Measurements

We measured the performance on our simulator using NETTALK[17] text-to-phoneme benchmark (as did Pomerleau et al.[15] and Blelloch&Rosenberg [3]). The network consists of an input layer with 203 units and a "true" unit, a hidden layer with 60 units, and an output layer of 26 units. The input layer is fully connected to the hidden layer, and the hidden layer is fully connected to the output layer. The true unit is fully connected to both the hidden and output layers. The total number of connections is 13,826. Our training set consisted of 12022 patterns. We updated weights once per epoch.

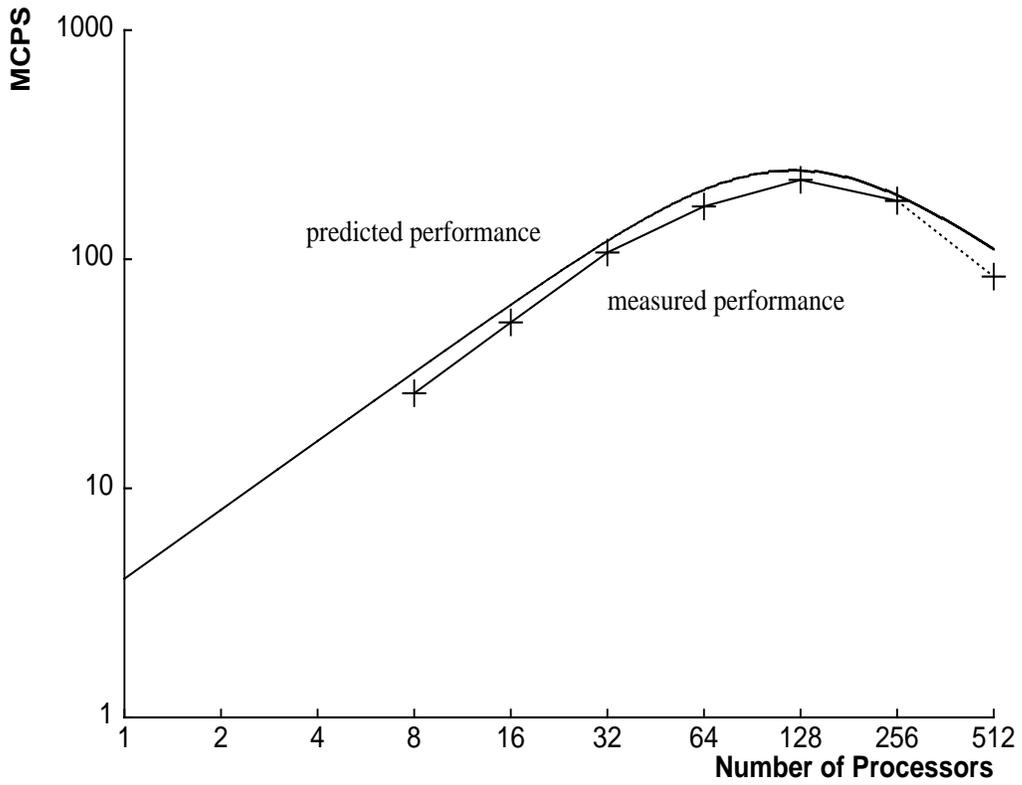

Figure 8: MCPS with Processors Configured in a Ring

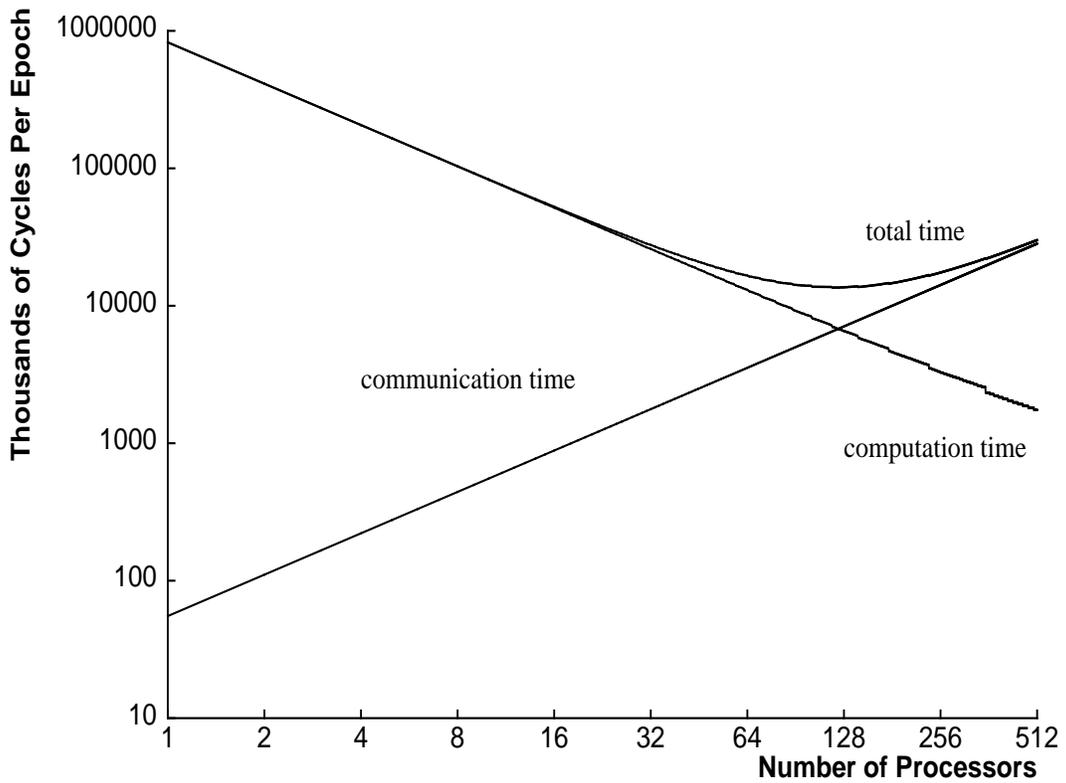

Figure 9: Communication and Computation with Processors Configured in a Ring

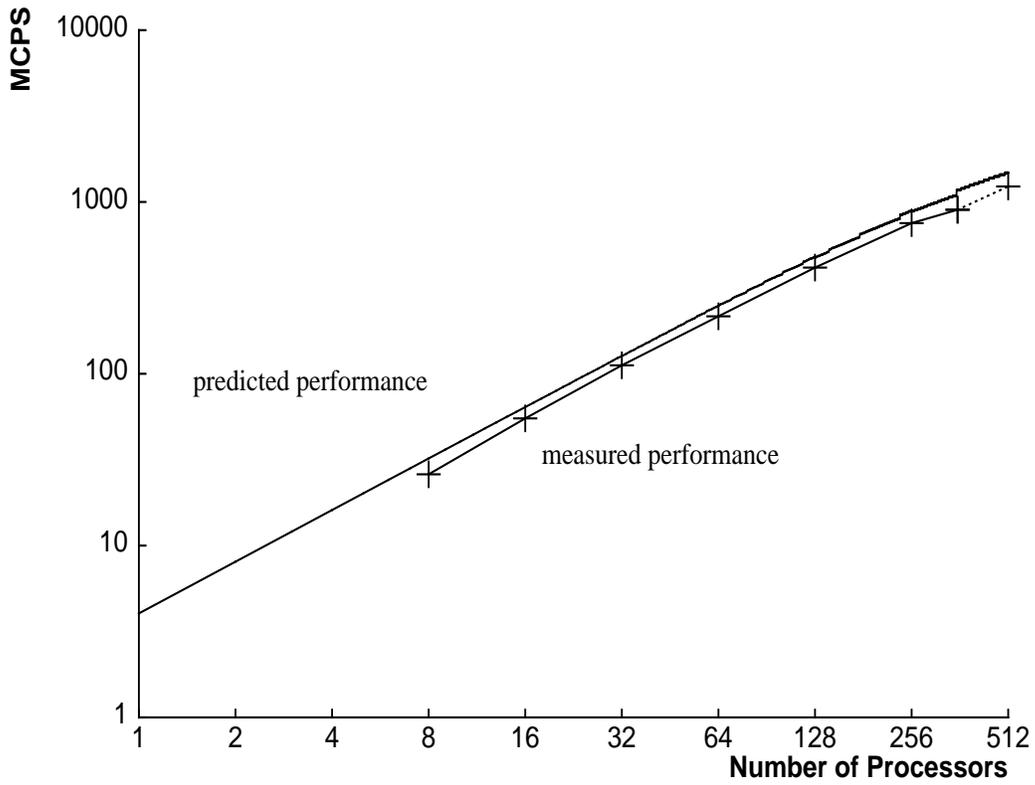

Figure 10: MCPS with Processors Configured in a Tree

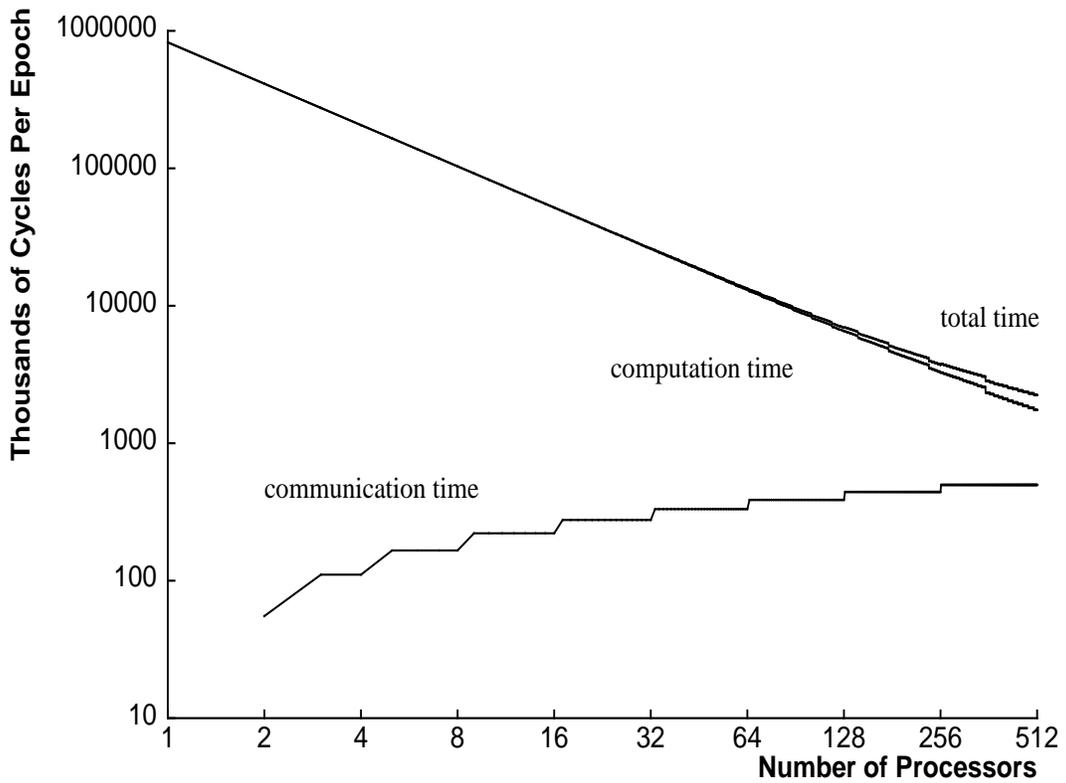

Figure 11: Communication and Computation with Processors Configured in a Tree

| Millions of Connections Per Second | | |
|---|---|---|
| **Processors** | **Tree MCPS** | **Ring MCPS** |
| 8 | 26 | 26 |
| 16 | 55 | 53 |
| 32 | 112 | 107 |
| 64 | 216 | 170 |
| 128 | 415 | 222 |
| 256 | 753 | 180 |
| 356 | 901 | — |
| 512 | 1231 | 84 |

Table 1: Millions of Connections Per Second, Summing in a Ring and in a Tree

As of August 1989, there were 356 fully operational processors for which the switch could be configured to provide, without errors, the interprocessor communication required by our application. The disk system was not operational. The timings with more processors are actual measurements, although the results generated by the program were not correct. Measurements do not include the time to initialize DRAM with training patterns. However, the measurements are real-time and therefore include all host and controller overheads.

Figures 8 and 10 show the predicted and measured performance of both types of summing. Figures 9 and 11 break down the execution time into components for communication and computation.

## 7. Comparison to Other Backprop Simulators

### 7.1. Difficulties with Making Comparisons

In this section, we will attempt to compare the performance of our backprop simulator on GF11 with the performance of simulators running on other machines. Before doing so, however, we believe it is important to note that while the simulators all implement essentially the same algorithm, there are variations between implementations, and between sets of test data which can make performance metrics somewhat misleading.

The most obvious source of difficulty in making these comparisons is that of variations on the algorithm caused by variations in hardware. Our simulator, for example, must pool updates over at least a number of I/O patterns equal to the number of operating processors to work at all, and four times that many to work efficiently. This hardware "limitation" (which is the source of our code's speed) means that we can't update weights frequently during a presentation of the NETTALK training set. Since the NETTALK training set is suited to frequent updates, our simulator would probably take more pattern presentations to learn this task than the Warp (with fewer processors) or the Connection Machine simulators (with a different form of parallelism). It is even conceivable that our simulator might take longer to learn NETTALK than one of these other simulators.

On the other hand, for the task we originally envisioned applying the simulator to, learning to do speech transcription by training recurrent networks over *huge* amounts of training data, the advantages of such frequent updates have not been demonstrated.

In short, using variations of an algorithm is sometimes useful, either for a particular task, or to extract a certain form of parallelism, but people reading performance comparisons should not be tempted to gloss over these variations; they may have significant effects on the utility of the simulator for some tasks.

One problem with measuring performance in connections per second is that the definition of this unit is not standard. Our definition [and Pomerleau's] is simply the number of connections (including connections to the true unit) times the number of patterns presented divided by the total time. Notice that this measure accounts for neither the frequency of weight updates nor the fact that connections from the input require less computation.

There is no clear solution to the problems of measuring performance. However, defining the performance metrics used and reporting the exact parameters of a benchmark make fair comparisons feasible. We have also found that our performance model elucidates performance differences between our simulator and other simulators.

### 7.2. Performance Data

Before we present our performance measurements, we must emphasize that our measurements do *not* include the I/O time to load training data onto processors. At the time these measurements were made, the GF11 disk system was not operational. See section 10.1 for a discussion of incorporating disk I/O into our simulator.

We should also emphasize that we updated weights once every *epoch*, while measurements for other machines include more frequent updates. We estimate that if we updated weights after every single pattern per processor (i.e., every 512 patterns), our peak rate would be dominated by communication time and DRAM transfer time. Using the performance model as a guide, the total time per update (on the NETTALK training set, using 512 processors) would be roughly ($\log_2 512$) communication steps, each taking four cycles per weight, and 20 cycles/weight of computation and memory transfers, yielding approximately 180 MCPS on 512 procs. This provides a lower bound on performance when more frequent updates are required. As an upper bound of theoretical interest, we estimate that as communication time approaches zero, (i.e., when updating once per epoch with huge data sets) the simulator would perform at approximately 1900 MCPS. We provide this figure only as an indication of the trade-off between communication and computation; it should not be considered a benchmark rating.

| Millions of Connections Per Second | |
|---|---|
| **Machine** | **MCPS** |
| $\mu$Vax | 0.008 |
| Sun 3/75 | 0.01 |
| Vax 780 | 0.027 |
| Sun 3/160/FPA | 0.034 |
| Ridge 32 | 0.05 |
| DEC 8600 | 0.06 |
| Convex C-1 | 1.8 |
| 16K CM-1 | 2.6 |
| Cray-2 | 7 |
| 64K CM | 13 |
| CM-2 | 40 |
| Warp | 20 |
| GF11 | 901 |

Table 2: Comparison of Learning Speeds for Various Machines

To convert MCPS to Megaflops, we consider only the computation on weights in the forward and backward passes, ignoring weight updates and the sigmoid calculation. Total computation time on weights depends on the fraction of weights that are in the input layer, i.e.,

$$\text{Megaflops } = MCPS \times (4 + 2\frac{W - W_i}{W})$$

For NETTALK, this factor is 4.23. Using our measurement for 356 processors, 901 MCPS is equivalent to 3.8 Gigaflops (of a possible 7.1 Gigaflops).

For purposes of comparison, table 7.2 gives performance figures for a number of other implementations of backpropagation learning. The Warp figure is from George Gusciora [personal communication]. The CM-2 figure is from Zhang et al.[18]. Other figures are from Blelloch and Rosenberg [3] or Pomerleau et al. [15]. Alexander Singer [personal communication] has an implementation for the CM-2, based on an algorithm suggested by Robert Farber, which is an order of magnitude faster than that of Zhang et al. The method used for reporting the performance of this simulation is not directly comparable with that used in this paper, but we estimate its performance at between 400 and 650 MCPS based on the figures we were given.

## 8. Fault Diagnosis and Fault Tolerance

The instability of the GF11 hardware during the code development period made fault diagnosis an essential part of our simulator.

To test the interconnection network, at program startup each processor receives initial values from the host and sends these values to over each switch configuration which the program will use during a run. The destination values are collected by the host and verified.

To test the processors, the simulator can be run in a diagnostic mode during which each processor executes backprop with the same training set and initial weights. Each processors acts as if it were the only processor; there is no inter-processor communication. The host then gets the total squared error for each processor, compares the values, and approves all processors that produce the most common total error.

Although separate microcode for computation and error detection is cleaner, it comes with a performance cost. We chose to integrate error detection with computation when there was a significant performance gain available by doing so. In the microcode routine that communicates weight changes, for example, each total weight change is tested for being in a "reasonable" range. If a value is out of range, it is set to zero and a fault count is incremented. This simple measure allows the program to work (i.e. reduce error) in the presence of processor and bus errors which would otherwise cause huge weights which saturate units and prevent further learning. Most of this error checking can be performed in the unused cycles during inter-processor communications.[11]

## 9. Limitations of the Simulator

Our approach to memory management restricts the size of the networks which may be simulated. The 512K words of DRAM hold training examples, weights, weight changes, and weight changes from the previous epoch. The maximum number of weights allowed is one-third of the available memory (512K minus the storage for training data). The maximum number of weights is also limited by microcode memory (512K lines). This latter limitation could be overcome if there are sets of groups of units with identical connectivity patterns. Then the address relocation mechanism can be used to map them onto the same section of microcode.

In SRAM, each unit requires one word for the net input to the unit (or Error on backward pass) and one word for the output of a unit. The lookup tables for the sigmoid calculation and other constants require approximately 1K words. The maximum number of units allowed is roughly $(16K - 1K - 2 \times maxbundle)/2$. This number is typically around 6000 units.

## 10. Extensions

### 10.1. Disk I/O

The GF11 disk system consists of 10 independent units which communicate with processors over the switch. (See figure 1). Each disk can deliver 8 Mbytes/sec for an aggregate peak rate of 80

---

[11]However, timings presented in this paper were taken with this part of the error checking turned off.

Mbytes/sec. At this rate, the entire DRAM of each processor can be filled in less than 15 seconds, and our NETTALK data set could be loaded in a fraction of a second.

Although we designed the simulator to load all training data at program startup, the disk system is fast enough to load patterns in parallel with training. The total capacity of the disk system, 4.5 Gbytes, is large enough to support very large applications, such as speech or image processing.

## 10.2. Switch Communication

Although it is not possible to sum $N$ numbers on $N$ processors in less than $O(\log_2 N)$ time, it is possible to perform $N$ such sums on $N$ processors in less than $O(N \log_2 N)$ time (See Figure 12). Intuitively, it takes $O(N)$ communications to sum $N$ numbers, so it should take $O(N^2)$ communications to sum $N$ sets of numbers if all communication steps can be fully utilized. Processors can pipeline the summing by sending partial sums of each weight change around a ring. On step $i$, processor $P$ receives a partial sum of *weightchange*$[(P + i + 1) \bmod N]$ from processor $(P + 1) \bmod N$ and adds its value for the element to its partial sum. After $N - 1$ steps, Processor $P$ has the sum of *weightchange*$[(P + N - 1) \bmod N]$. The sums are then circulated using another $N - 1$ steps.

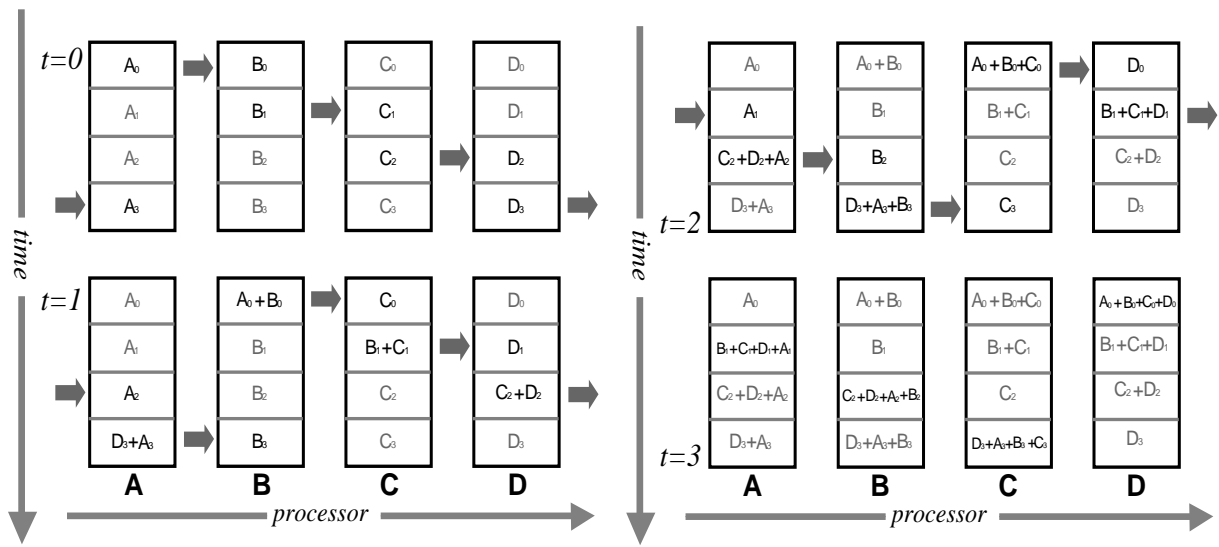

Figure 12: Pipelined Summing with Processors Configured in a Ring

In practice, the constant factor for this algorithm on GF11 is not significantly better than the log of the number of processors because of the processor dependent computation. But this extension is important in theory because it scales better with the number of processors; it indicates that a very simple processor interconnection, a ring, is adequate for our approach to parallelizing backprop provided that processor dependent addressing is possible.

## 10.3.  Improving Performance for Small Networks

In general, our code is optimized for networks with large enough layers to fill up the 25 cycle pipeline. However, some neural network architectures for large applications have small layers, e.g., classification tasks with a few output units[10].

Speeding up computation on smaller layers requires minimizing the longest chain of dependent operations so that more operations can be interleaved. In the forward pass, 2 floating point operations per weight are executed. Multiplies are independent of each other, but all the additions to the net input of a destination form a linear dependency chain. This chain can be shortened by rearranging the operations so that the dependency graph forms a binary tree. To sum an array $x$ of size $N$ (assuming $N$ is a power of 2 for simplicity),

```
for i = 0 step 1 to lg(n)-1
  for j = 0 step (2^(i+1)) to n-1
        x[j]=x[j]+x[j+2^i];
```

The result is left in x[0].

## 10.4.  Fault Tolerance

While we were developing the simulator described in this report, an effort was being made to rapidly increase the number of available processors in GF11. This meant that the hardware was relatively unstable, a situation which lead us to consider ways how the program could be made more robust. Although the machine is far more stable at the time of writing [David George,[12] personal communication] than it was during program development, there are a number of interesting measures which could make the program more fault tolerant.

Since we contemplate storing huge amounts of training data in RAM (i.e., up to 1.1 GBytes) there is some possibility of changes in the stored data due to transient memory errors. In order to detect such occurrences, we would like to add checksums to the training data. Such checksums could be computed very rapidly at infrequent intervals during training. As well as facilitating fault detection, checksums would provide additional utility by allowing the program to decide whether training patterns need to be reloaded following runs of other users programs.[13] Pending a fully functioning disk I/O system, such a facility could be very useful.

The large amount of memory on each processor enables us to contemplate another form of fault tolerance: weight checkpointing to memory. Every few epochs, a copy of the networks weights could be copied to one of a number of checkpoint spaces in each processor's RAM (or trading time against space, spread across the RAM of all processors over the switch). In the event that an error in learning is detected (e.g. by a large increase in total pattern error between epochs), one of the checkpoints in RAM could be copied back as active weights and learning restarted. This technique should make learning virtually immune to transient or intermittent hardware failures.

---

[12]David George is the manager of the GF11 project at IBM.

[13]This is not always necessary, the main switch diagnostic program for GF11, for example, only clobbers memory below that used for storing training cases by our program.

One of the problems with distributing weight changes, rather than weights, to all processors during the update cycle is that the weights may begin to take on different values on different processors. There are two ways in which this could occur. First, an arithmetic or memory failure on a processor could change a weight. Secondly, and less obviously, the order in which weight changes are added together is different on different processors. Since the numbers are floating point with a fixed representation, this can result in different totals on different processors, i.e., floating point operations are not necessarily associative. Small differences between weights on different processors can probably be tolerated over a few training epochs, but they need to be prevented from accumulating. We propose to ensure that all processor work on more or less the same network by occasionally averaging and rebroadcasting the "reasonable" values for each weight across processors.

## 10.5.    Recurrent Networks

The final goal of this work was applying backpropagation training to the problem of speech recognition using recurrent (Elman) nets. The important difference between these networks and those with a conventional, feed forward structure is that unit activation levels are copied to units in *earlier* layers between presentations of successive patterns in a sequence. Since a unit's previous value contributes to its input, the network can learn weights that allow it to retain useful information over time.

The copying of activations between pattern presentations presents a slight difficulty for parallelization. Activations must be copied from units in the copy of the network which saw the previous Input/Output pattern in a sequence. This means that if we spread cases from sequences across processors we would have to do the costly operation of communicating activation levels across the switch. The secret in avoiding this is to repeat the method used in Witbrock's previous vectorized Elman net simulator for the Convex C-1: parallelizing across sequences instead of across cases. This means that all copying of activation happens within the processor responsible for a particular sequence. While the actual copying of activations is trivial — a small piece of microcode is constructed from a list of *(from unit, to unit)* pairs at microcode generation time — parallelizing across cases involves a couple of extra complications. First, a processor must know when a new sequence starts within its patterns, so that it can zero out the units which receive copies of activations. This is done by keeping a flag with each pattern: 1.0 if it's the continuation of an old sequence and 0.0 if it's the start of a new sequence, and multiplying the copied activations by this flag before copying into the destination units. The second change is that one cannot rely on all processors having the same number of patterns, since sequences may differ in length. Any patterns beyond the end of the last sequence on each processor must be marked with a flag indicating that no error should be back-propagated for this pattern. Again, this is done by multiplying the error by the value of this flag.

## 11.   Evaluation of Architectural Features

In building our simulator and writing this report, we have had two equally important goals. The first goal was produce the fastest simulator possible on the hardware while maintaining enough flexibility to make it useful for our other research purposes. For this goal the chief measure of success is raw speed — millions of connections per second. The other goal was to evaluate *features* of an architecture which make it suitable or unsuitable for simulating connectionist networks, not just the performance of the whole package.

We begin by analyzing alternative strategies for communication. The performance of the ring summing algorithm indicates the limitations of a single ring of processors with no processor-dependent addressing. Clearly, figure 8 indicates that this approach does not scale well with the number of processors. The Beneš Network allows more sophisticated communication which significantly improves performance. However, the pipelined ring summing algorithm presented in section 10.2 has the best asymptotic complexity. Given enough weights, the total time is independent of the number of processors. We conclude that for our approach to parallelizing backprop, a sophisticated communications network is not necessary, if processor-dependent addressing is available.

Although SIMD control appears at first glance to be a limitation, we did not find any need for more complex control. In fact, we found that the single thread of control made code easy to design and debug. Moreover, the GF11 processor architecture matched our needs quite well. The balance of adds and multiplies inherent in the backprop algorithm allowed us to keep the floating point units busy, and the table lookup facility proved invaluable for the sigmoid calculation.

However, the three-level, software-controlled memory hierarchy added to the program complexity more than any other feature of the machine.[14] How could this complexity be avoided? It could *not* be prevented by a hardware controlled cache because of the pattern of the sequential accesses to large arrays. It is uncertain that compiler support would help either. Simply reducing the memory capacity in order to provide a uniform access time would place unacceptable constraints on simulation size. The best solution would be to increase the number of interleaved memory banks. It is relatively easy to arrange for weights to be accessed strictly sequentially. The bandwidth would have to be increased, but the latency would not have to be decreased.

In fact we found a general principle for simulating large networks: bandwidth is much more important that latency. This principle applies to computation on weights, memory accesses to weights, and inter-processor communication. Even though the processor pipeline is 25 deep, our code was naturally amenable to pipeline scheduling without any deliberate restructuring.

---

[14]We were not alone in finding this; Yurij Baransky [personal communication] reports that the memory hierarchy has presented difficult design decisions when mapping other applications to GF11.

## 12.  Conclusions

This paper reports our experience in mapping the backprop algorithm onto the GF11. We feel that there are several contributions that this work makes to the enterprise of neural network simulation:

We formalized the structure of the storage access patterns during the algorithm and used this model to design an efficient mapping on to a software controlled 3-level memory hierarchy. Although the fine details of this mapping are specific to GF11, the technique of accessing a subset of weights over several training examples is directly applicable to other machines, even machines with a hardware controlled cache.

We measured the simulator performance empirically, and more importantly produced an analytical model of its performance. This allowed us to independently evaluate the contribution to performance on this task of the various architectural features of GF11.

Finally, we showed that although GF11 was conceived with a particular application in Quantum Chromodynamics[1] in mind, the architecture is also well-suited to other applications.

## 13.  Acknowledgements

We would like to thank Scott Fahlman (CMU), Danny Sabbah (IBM), and Dave George (IBM) for initiating this project. We are very grateful to Dave George, Yurij Baransky, Jim Sexton, and Micky Tsao at IBM for their technical assistance and helpful advice, and to Michael Portuesi, Micky Tsao, and Michael Franzini for help shipping data and equipment between CMU and T.J. Watson. And we also thank the CMU Boltzmann group, Yurij Baransky, Guy Blelloch, Scott Fahlman, Alan Fisher, and Dean Pomerleau for their helpful comments on the presentation of this work.